\documentclass[11pt]{article}
\pdfoutput=1

\usepackage[margin=1in]{geometry}
\usepackage[T1]{fontenc}
\usepackage[utf8]{inputenc}
\usepackage{lmodern}
\usepackage{graphicx}
\usepackage{array}
\usepackage{longtable}
\usepackage{booktabs}
\usepackage{threeparttable}
\usepackage{url}
\usepackage{float}
\usepackage{listings}
\usepackage{changepage}

\usepackage{seqsplit}
\let\oldtexttt\texttt
\renewcommand{\texttt}[1]{\oldtexttt{\protect\seqsplit{#1}}}

\usepackage{sectsty}\sectionfont{\large\bfseries}\subsectionfont{\normalsize\bfseries}\subsubsectionfont{\normalsize\itshape}

\begin{document}


\begin{center}
    {\Large \textbf{A large-scale corpus of religious radio broadcast transcripts from webstream recordings in the United States}}\\[1.2em]


    {\large
    Samuel Bestvater$^{1,*}$,
    Athena Chapekis$^{1}$,
    Skyler Seets$^{1}$,
    Anna Lieb$^{1}$,
    Sono Shah$^{1}$,
    Aaron Smith$^{1}$\\[0.7em]
    }

    {\normalsize
    $^{1}$Pew Research Center, Washington DC, USA\\[0.5em]
    *Correspondence: labsinfo@pewresearch.org
    }
\end{center}

\begin{abstract}
\begin{adjustwidth}{0.35in}{0.35in}
Religious radio is a widespread but understudied form of mass communication in the United States, and content-level analysis of it has been constrained by the absence of large-scale transcript data.
This Data Descriptor presents a corpus of transcribed English-language religious radio broadcasts captured from live webstreams over a one-month period in July 2025.
Fifteen-minute segments were recorded on a rolling schedule from 785 distinct streams, which together rebroadcast the signals of more than two thousand AM and FM stations, yielding over 700,000 recordings and more than 60 million diarized transcript lines.
Each recording was transcribed and speaker-diarized with an automated pipeline, and segmented and labeled by programming format and topic using a large language model.
The corpus is organized as linked tables of stream metadata, recording metadata, and transcript lines.
It supports descriptive study of religious broadcasting across regions and traditions, analysis of how social and political issues are discussed in religious media, and speech-processing research in an underrepresented domain.
\end{adjustwidth}
\end{abstract}

\section*{Background \& Summary}

Faith-based radio programming has a long history in the United States \cite{hangen2002} and remains an enduring and widespread form of mass communication.
According to a recent nationally representative survey, 45\% of U.S. adults today report that they listen to at least some form of religious audio programming \cite{bestvater2026religiousradio}.
The audience for this content cuts across religious and political lines, encompassing large majorities of White evangelical (76\%) and Black Protestants (84\%), roughly four-in-ten Catholics, and even 18\% of those with no religious affiliation.
Broadcast radio remains the most common means of access, with roughly three-quarters of religious audio consumers tuning into terrestrial AM/FM stations. 

The prominence of religious programming in American radio is the result of a considerable institutional history.
Religious broadcasters were among the earliest adopters of the medium, and over the course of the twentieth century, evangelical Protestants in particular came to occupy a dominant position on the airwaves, aided by shifts in broadcasting regulations and the economics of paid religious programming \cite{schultze1988, voskuil1990}.
Scholars of media and religion have long treated the ``electronic church'' as a significant object of study, examining its audiences, institutional forms, and role in the broader religious lives of Americans \cite{hoover1988, horsfield1984}.
Beyond its spiritual function, religious broadcasting has also served as an infrastructure for political communication, and has been associated with the coherence and mobilization of religious constituencies \cite{calfano2021}.

Despite its historical and contemporary significance, systematic analysis of the content of religious radio programming has been limited by the absence of large-scale, content-level data \cite{mcdonnell2023}.
Prior empirical work has largely relied on listener surveys, studies of station operations and institutional strategy, or close analysis of a small number of programs or stations \cite{greer2003}.
Large-scale, content-level data have been comparatively scarce: existing broadcast speech corpora have tended to focus on news or general talk radio rather than religious programming, and have typically been limited in the number of stations or the breadth of formats they capture \cite{beeferman2019}. 
As a result, basic descriptive questions about what is broadcast, as well as how programming varies across regions, formats, and religious traditions have been difficult to address empirically across the medium as a whole.  

To enable content-level study of religious radio at scale, we introduce the Religious Radio Corpus: a large-scale corpus of transcribed religious radio broadcasts in the United States.
Over the course of July 2025, we monitored 785 distinct live webstreams, of which 779 are represented in the released corpus (see Data Records).
These streams collectively rebroadcast the signal of over 2,000 religious AM and FM radio stations across the country. We captured 15-minute recordings from each stream on a rolling basis, yielding 715,688 distinct recordings which together make up over 60 million transcribed lines of speech.
Each recording was transcribed and speaker-diarized using an automated pipeline \cite{bain2023whisperx, plaquet2023powerset, bredin2023pyannote}. 
The corpus is structured at three levels: stream- and station-level metadata, recording-level metadata, and the diarized transcript lines themselves.
Together, the corpus captures programming from nearly all religious radio stations in the United States that broadcast an accessible webstream.

The Religious Radio Corpus is designed to support research across several fields.
For scholars of religion and media, it enables descriptive study of what is broadcast on religious radio--the balance of music, sermons, and talk programming, and how it varies across regions and religious traditions.
For political communication and the social sciences more broadly, the transcripts and topic annotations support analysis of how political and social issues are discussed within religious media. 
For researchers in speech and language processing, the corpus provides a large collection of naturalistic, multi-speaker transcribed speech in a domain underrepresented in existing corpora, supporting work on domain adaptation, topic classification, and the modeling of multi-speaker interaction.
More broadly, the geographic and denominational breadth of the captured stations makes it possible to study religious broadcasting as a national phenomenon, rather than through individual stations or programs.

These data were previously used in a Pew Research Center report on religious radio in the United States \cite{bestvater2026religiousradio}, which combined the broadcast corpus with a nationally representative survey and station-level administrative data to describe the religious radio landscape--including the geographic distribution of stations, the balance of music and spoken programming, and the prevalence of political and social content. 
That report presented aggregate findings; the present corpus releases the underlying transcripts and recording-level metadata in full, enabling the kinds of fine-grained, record-level analysis described above. 
The most closely related prior dataset is RadioTalk \cite{beeferman2019}, a corpus of transcribed U.S. talk radio from a sample of stations; the Religious Radio Corpus differs in its focus on religious programming specifically, its near-complete coverage of web-accessible stations within that domain, and its accompanying station- and recording-level metadata. 
More broadly, the Religious Radio Corpus joins a body of large-scale spoken-language corpora, including resources for broadcast news—from transcribed-speech corpora \cite{graff1997hub4ldc, graff1997hub4trans} to large-scale television-news collections \cite{internetarchive_tv}—along with corpora of parliamentary proceedings \cite{wang2021voxpopuli} and podcasts \cite{clifton2020spotify}.
It also builds on foundational corpora of conversational speech such as Switchboard \cite{godfrey1992switchboardpaper, godfrey1993switchboard}, Fisher \cite{cieri2004fisherpaper, cieri2004fisher}, CALLHOME \cite{ldc1997callhome}, and the Santa Barbara Corpus of Spoken American English \cite{dubois2005sbcsae}.
It is distinguished within this landscape less by its scale than by its focus on a single, comprehensively covered broadcast domain.

\section*{Methods}

Data collection for the Religious Radio Corpus involved an end-to-end pipeline consisting of station selection, audio capture, transcription, diarization, and metadata annotation.
(See Fig. 1 for an overview.)
Each stage is described in detail below.

\begin{figure}[H]
    \centering
        \caption{Data collection and processing pipeline}
    \includegraphics[width=\linewidth]{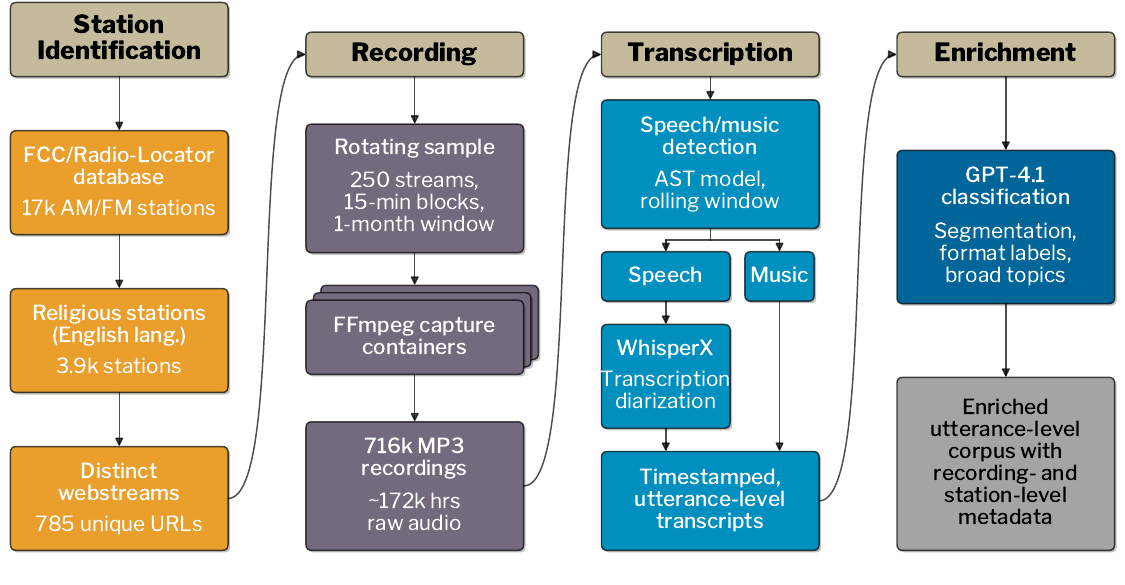}
\end{figure}

\subsection*{Study design and population definition}

The sampling frame for this dataset was derived from a March 2025 snapshot of the Radio-Locator database \cite{radiolocator}, a commercially maintained registry of FCC-licensed terrestrial radio stations that augments administrative licensing data with station-level metadata including call sign, band and frequency, geographic coverage area, primary programming genre, website URL, and live audio stream URL.
The full snapshot consists of 25,753 stations located across the 50 U.S. states and the District of Columbia, of which 17,115 were primary stations broadcasting on the AM or FM band; the remaining 8,638 were secondary booster or translator stations that rebroadcast a primary signal.

Religious stations were identified by filtering on Radio-Locator's ``format'' field, a mutually-exclusive label describing the primary genre of each station's programming. 
Stations were considered to be religious stations if their primary genre was listed as ``Religious,'' ``Christian Contemporary,'' ``Gospel Music,'' or ``Spanish Christian,'' yielding 4,328 primary AM and FM stations, with another 3,038 booster or translator stations.
A total of 424 primary and 152 secondary stations classified as Spanish Christian were subsequently excluded from the sample frame due to limitations in the automated transcription and classification pipeline's ability to process non-English audio.
The remaining population of English-language religious radio stations consisted of 3,904 primary stations, with another 2,886 booster and translator stations rebroadcasting their signal.

Not all stations in the population had a publicly accessible live audio stream URL available at the time of the data collection.
Restricting to stations that did yielded 2,083 primary stations with a retrievable stream URL (as well as another 1,889 booster and translator stations) that could be included in the recording protocol.
Because it is common practice for stations under common ownership to share a single broadcast feed, these 2,083 stations resolved to 785 distinct stream URLs, which constituted the operational recording population.
Content recorded from each unique stream URL can be attributed to all stations sharing that URL.

\subsection*{Audio capture}

Audio content was collected from the 785 distinct stream URLs identified in the population definition stage over a one-month recording window spanning July 2025.
Rather than continuously recording a subset of streams, the protocol was designed to maximize coverage across all stations by sampling uniformly in time: 250 recording slots were scheduled every 15 minutes, 24 hours a day, for the full 31-day collection window, yielding a theoretical maximum of 744,000 scheduled recording attempts.
Streams were rotated across these slots such that each unique URL was allocated roughly 240 hours of scheduled recording time over the course of the month.

To execute this protocol at scale, we developed and deployed a cluster of 250 containerized stream listener applications, each operating according to a dedicated recording schedule.
At the start of each 15-minute slot, the assigned listener connected to its scheduled stream URL and captured the incoming audio using FFmpeg v4.1 \cite{ffmpeg}.\footnote{Recording windows were exactly 15 minutes in duration, but the actual captured audio was typically slightly shorter due to the time required to spin up the container and establish a connection to the stream.}
After 15 minutes, the listener disconnected and saved the recording as a low-bitrate (64 kbps) MP3 file. Of the 744,000 scheduled recording attempts, 716,436 produced a captured audio file, of which 715,688 were retained after processing (see Technical Validation), a 96.2\% end-to-end yield.
The resulting corpus contains transcriptions of approximately 172,000 hours of raw audio across the 785 unique stream URLs.\footnote{Because multiple primary stations frequently share a single stream URL, the corpus is released at the level of unique stream URLs rather than individual stations.
Users can expand the corpus to the station level by joining on the station-to-stream mapping table included in the data release, which identifies all stations associated with each unique URL.}

\subsection*{Transcription \& diarization}

Prior to transcription, each 15-minute MP3 recording was passed through a speech/music classification step to partition the audio into contiguous regions of speech and music.
This was implemented using MIT/ast-finetuned-audioset-10-10-0.4593 \cite{ast_audioset}, a pretrained Audio Spectrogram Transformer (AST) model fine-tuned on the AudioSet dataset \cite{gemmeke2017audioset}.
To apply this clip-level classifier to 15-minute recordings, each file was chunked into overlapping 10-second segments, each of which was independently classified as speech or music.
Contiguous classifications of the same type were then merged back into continuous regions.
In cases where the model assigned both labels to a segment--likely reflecting instances where speech overlapped a musical background--the segment was designated as speech.
Regions classified as music were carried forward into the transcript as \texttt{<MUSIC>} placeholder tokens rather than being passed to the transcription model.

Speech regions were then processed through the WhisperX transcription system \cite{bain2023whisperx}, which combines an automatic speech recognition model with models for speaker diarization and voice activity detection.
WhisperX extends standard speech-to-text output with word-level timestamp alignment and speaker differentiation, producing utterance-level transcripts in which each line represents a single speaker turn annotated with a start time, end time, and anonymized speaker identifier.
Our implementation of the WhisperX pipeline utilized OpenAI's whisper-large-v3-turbo model \cite{radford2023whisper, whisper_large_v3_turbo} for ASR and pyannote's speaker-diarization-3.1 model for speaker diarization and voice activity detection \cite{bredin2023pyannote, pyannote_model}.
Performance metrics for this system are reported in the Technical Validation section. 

\subsection*{LLM-based metadata annotation}

Each recording-level transcript was passed to OpenAI's GPT-4.1 API endpoint \cite{openai2025gpt41} to perform two annotation tasks: segmentation and classification.
In the first step, the model partitioned each transcript into topically coherent segments by identifying natural boundaries in subject matter, interactional format, and program structure--for example, transitions between advertisements, sermons, and host-led discussion.
For each resulting segment, the model assigned a single mutually exclusive format label drawn from a fixed codebook, and one or more topic labels from a separate codebook permitting multiple simultaneous assignments.
The model prompt used for this annotation process as well as the full codebooks for format and topic classification are included in the appendix.

Format labels capture the communicative function of a segment and contain the following categories: discussion/monologue/commentary, interview, caller interaction/audience participation, news read/traffic/ weather, religious service/sermon, audio drama/narrative, ad/promotion, and transition/filler.
Topic labels capture the broad subject matter of a segment and can contain: religion, politics/current events/social commentary, lifestyle/advice/personal development, family/parenting/education, health/wellness, entertainment/pop culture/sports, business/economics/finance, and science/technology.
Both codebooks were developed through an iterative human coding process prior to model deployment, and model performance was validated against human-coded samples; accuracy metrics for both classifiers are reported in the Technical Validation section.

The dataset includes two array-valued fields per recording — \texttt{predicted\_programming\_formats} and \texttt{predicted\_topic\_labels} — each containing the set of distinct labels observed across all segments within that recording.


\section*{Data Records}
The Religious Radio Corpus is released through a dual-repository model, with the complete corpus deposited in both repositories.
An archival version of record is deposited in Harvard Dataverse \cite{religiousradio-dataverse}, and an access-optimized version is available through Hugging Face Datasets \cite{religiousradio-huggingface}.
Both repositories contain the same core data, but are organized differently to support different use cases.
The Harvard Dataverse deposit constitutes the citable release and, alongside the data tables, includes the schema and codebook files, the LLM annotation prompt, a release manifest recording per-file checksums and row counts, and full release documentation; it is intended for long-term preservation and citation.
The Hugging Face repository provides the same data in analysis-ready Apache Parquet format, accompanied by a dataset card, and is intended as the primary layer for streaming, filtering, and large-scale computational reuse.
The description below refers primarily to the Parquet organization of the Hugging Face release, although the identical tables are present in the Dataverse deposit.

The corpus is organized as three linked tables---\texttt{stations}, \texttt{recordings}, and \texttt{transcript\_lines}---that correspond to the three levels at which the data are structured (monitored stream, 15-minute recording, and diarized transcript line).
These tables can be joined using stable identifiers: \texttt{recordings.station\_id} is a foreign key to \texttt{stations.station\_id}, and \texttt{transcript\_lines.recording\_id} is a foreign key to \texttt{recordings.recording\_id}.
A crosswalk table (\texttt{station\_stream\_crosswalk.parquet}) maps each monitored stream to all individual stations that carry its signal, supporting expansion of stream-level content to the station level.

Both repositories are organized as follows:

\begin{verbatim}
stations/stations.parquet
recordings/recordings.parquet
crosswalk/station_stream_crosswalk.parquet
transcript_lines/YYYY-MM-DD/part-*.parquet
README.md
\end{verbatim}

The \texttt{stations} and \texttt{recordings} tables are each stored as a single Parquet file.
The \texttt{transcript\_lines} table is partitioned by broadcast date, with one subdirectory per day of the collection window (\texttt{2025-07-01} through \texttt{2025-08-01}) and one or more Parquet part files within each.\footnote{These day-level partitions reflect the UTC date of the recording start time, which may differ from the station's local date due to timezone offsets. Recording ran from 2025-07-01 00:00:00 EDT through 2025-07-31 23:59:59 EDT, which corresponds to 2025-07-01 04:00:00 UTC through 2025-08-01 03:59:59 UTC.}
This partitioning allows data from a single day, or a range of days, to be loaded or streamed without reading the full corpus.

The \texttt{stations} table contains one row per monitored stream URL (the operational recording unit), with the fields defined in Table~\ref{tab:stations}.

\begin{table}[H]
\centering
\caption{Fields in the \texttt{stations} table.}
\label{tab:stations}
\small
\begin{tabular}{p{1.25in} p{0.75in} p{4in}}
\toprule
Field & Type & Description \\
\midrule
\texttt{station\_id} & string & Primary key; unique identifier for the stream entry \\
\texttt{callsign} & string & Call sign of the station associated with the stream (e.g., WBNH-FM) \\
\texttt{band} & string & Frequency band: \texttt{AM} or \texttt{FM} \\
\texttt{region} & string & U.S. Census region: Northeast, South, Midwest, or West \\
\texttt{timezone} & string & Station's local timezone in IANA format (e.g., America/Chicago) \\
\texttt{utc\_adjust} & float & Offset in hours between the station's local time and UTC \\
\texttt{related\_stations} & list[string] & Identifiers of all stations that share this stream URL \\
\bottomrule
\end{tabular}
\end{table}

Although 785 distinct stream URLs were included in the recording protocol (see Methods), the released \texttt{stations} table contains 779 rows.
Five stream URLs yielded no usable audio over the entire collection window and are excluded, and one additional station was removed after review: although listed as religious in the March 2025 Radio-Locator snapshot, it had changed to a classic hits format by the time data collection started and was not observed to broadcast religious content during the recording window.

Because stations under common ownership frequently share a single broadcast feed, a stream entry may correspond to more than one licensed station; \texttt{related\_stations} records the full set of stations attributable to that stream, and content recorded from a stream can be attributed to all of them. 
This information is repeated for more convenient access in the \texttt{crosswalk/station\_stream\_crosswalk.parquet} table, which contains one row per station-stream pair, with the fields defined in Table~\ref{tab:crosswalk}. 

\begin{table}[H]
\centering
\caption{Fields in the \texttt{station\_stream\_crosswalk} table.}
\label{tab:crosswalk}
\small
\begin{tabular}{p{1.25in} p{0.75in} p{4in}}
\toprule
Field & Type & Description \\
\midrule
\texttt{station\_id} & string & Foreign key to \texttt{stations.station\_id} (the monitored stream entry) \\
\texttt{stream\_station\_id} & string & Identifier of an individual station carried by the stream \\
\texttt{is\_primary} & boolean & True for primary stations (AM, FM, and low-power FM); false for boosters and translators \\
\bottomrule
\end{tabular}
\end{table}

The \texttt{recordings} table contains one row per captured 15-minute recording, with the fields defined in Table~\ref{tab:recordings}.

\begin{table}[H]
\centering
\caption{Fields in the \texttt{recordings} table.}
\label{tab:recordings}
\small
\begin{tabular}{p{1.75in} p{0.75in} p{3.5in}}
\toprule
Field & Type & Description \\
\midrule
\texttt{recording\_id} & integer & Primary key; unique identifier for the recording \\
\texttt{station\_id} & string & Foreign key to \texttt{stations.station\_id} \\
\texttt{day} & string & Broadcast date of the recording (YYYY-MM-DD) \\
\texttt{recording\_start\_utc} & timestamp & Start time of the recording in UTC (ISO 8601) \\
\texttt{recording\_duration\_sec} & float & Duration of the recording in seconds \\
\texttt{predicted\_programming\_formats} & list[string] & Distinct programming-format labels across the recording's segments \\
\texttt{predicted\_topic\_labels} & list[string] & Distinct topic labels across the recording's segments \\
\bottomrule
\end{tabular}
\end{table}

The \texttt{predicted\_programming\_formats} and \texttt{predicted\_topic\_labels} fields hold the union of the labels assigned to the recording's constituent segments during LLM-based annotation (see Methods); the underlying codebooks are given
in~Tables~\ref{tab:format-codebook} and~\ref{tab:topic-codebook} in the Appendix.
\texttt{predicted\_topic\_labels} may be empty for recordings in which no segment received a topic label.

The \texttt{transcript\_lines} table contains one row per diarized utterance (or music region), with the fields defined in Table~\ref{tab:transcript_lines}.

\begin{table}[H]
\centering
\caption{Fields in the \texttt{transcript\_lines} table.}
\label{tab:transcript_lines}
\small
\begin{tabular}{p{1.25in} p{0.75in} p{4in}}
\toprule
Field & Type & Description \\
\midrule
\texttt{recording\_id} & integer & Foreign key to \texttt{recordings.recording\_id} \\
\texttt{station\_id} & string & Identifier of the associated station (denormalized for convenience) \\
\texttt{line\_index} & integer & Sequential index of the line within its recording (0-based) \\
\texttt{start\_sec} & float & Start time of the line, in seconds from the recording start \\
\texttt{end\_sec} & float & End time of the line, in seconds from the recording start \\
\texttt{duration\_sec} & float & Duration of the line in seconds \\
\texttt{speaker\_id\_local} & string & Speaker label, unique only within a recording; \texttt{<MUSIC>} marks a music region \\
\texttt{text} & string & Transcribed text of the utterance, or \texttt{<MUSIC>} for a music region \\
\texttt{is\_music} & boolean & \texttt{True} if the segment was classified as music by the audio classifier \\
\bottomrule
\end{tabular}
\end{table}

Speaker labels take the form \texttt{SPK\_00}, \texttt{SPK\_01}, and so on.
They are assigned independently within each recording and are not consistent across recordings: \texttt{SPK\_00} in one recording does not denote the same person as \texttt{SPK\_00} in another.
Lines with \texttt{is\_music~=~True} carry the placeholder \texttt{<MUSIC>} in both \texttt{speaker\_id\_local} and \texttt{text} in place of transcribed speech.

\section*{Technical Validation}

\subsection*{Data coverage and completeness}

The Religious Radio Corpus was produced by rotating 250 continuous stream recorders over available streams for a one-month period. Of 744,000 scheduled 15-minute recording slots over the collection window, 716,436 (96.3\%) produced a successfully captured audio file; network and stream outages account for the remainder. Of these, 715,688 recordings (96.2\% of scheduled slots) are retained in the released corpus after excluding a small number of files that failed transcription or downstream segmentation, recordings that could not be matched to a scheduled slot, and one station whose content did not meet the study's inclusion criteria (Table~\ref{tab:coverage}).

\begin{table}[H]
\centering
\caption{Data collection and processing coverage}
\label{tab:coverage}
\begin{threeparttable}
    \small
\begin{tabular}{lrr}
\toprule
Pipeline stage & Recordings & \% of scheduled \\
\midrule
Scheduled recording attempts & 744,000 & 100.0 \\
Successfully captured & 716,436 & 96.3 \\
Retained after processing and content filtering & 715,688 & 96.2 \\
\bottomrule
\end{tabular}
\begin{tablenotes}
\footnotesize
\item Notes: Coverage of the recording and processing pipeline over the
one-month collection window. ``Scheduled recording attempts'' is the theoretical
maximum of 250 concurrent recording slots $\times$ 96 fifteen-minute slots per day
$\times$ 31 days. ``Successfully captured'' counts recording slots that produced a
raw audio file. ``Retained after processing and content filtering'' counts recordings
present in the released \texttt{recordings} table, after removing recordings that
failed transcription or segmentation, recordings that could not be matched to a
scheduled slot, and recordings from one station excluded as non-religious (see text).
The final row corresponds to an end-to-end pipeline yield of 96.2\%.
\end{tablenotes}
\end{threeparttable}
\end{table}

Capture failures were stable over time but unevenly distributed across stations.
Daily failure counts remained low and roughly constant across the collection window (mean $\approx$ 883 failed slots per day, range 692--1,339), with no sustained interruptions, indicating that missed recordings do not correspond to a systematic gap in coverage of any part of the month.
Across stations, however, failures were concentrated in a small number of unreliable streams: beyond the five streams that
yielded no usable audio at all (excluded from the release; see the Data Records section), seven additional streams were captured at fewer than half of their scheduled slots.
The remaining streams were captured at consistently high rates.

\subsection*{Transcription quality}

To assess the accuracy of the automated transcription pipeline, we compared ASR-generated transcripts against human reference transcriptions on a stratified validation sample of 600 recording clips drawn from the corpus---9.26 hours of audio in total.\footnote{One clip was excluded from the final analysis because its reference contained no transcribable speech, leaving 599 clips for this comparison.}
To establish a human performance baseline, a subset of 100 clips was independently transcribed a second time, allowing an inter-annotator (Human--Human) comparison on the same footing as the Human--ASR comparison.

Word error rate (WER) was computed using the \texttt{jiwer} library \cite{jiwer}.
Prior to alignment, both reference and hypothesis transcripts were normalized by lowercasing, removing punctuation, collapsing repeated whitespace, and converting written-out English numbers to their numeric form (for example, ``twenty twenty-five'' to ``2025''), so that superficial formatting differences are not counted as errors.
WER is reported as the sum of substitutions, deletions, and insertions divided by the number of reference words, and is aggregated within each reported group by pooling error counts and reference lengths across clips (a duration-weighted rather than clip-averaged WER).
Substitution, deletion, and insertion rates are reported separately as each error type divided by reference length. Confidence intervals were obtained by bootstrap resampling of clips within each group (1{,}000 resamples).

Overall, the ASR pipeline achieved a WER of 5.14\% against human reference transcriptions, compared with a Human--Human inter-annotator WER of 2.57\% on the double-transcribed subset (Table~\ref{tab:wer-by-stratum}).
The gap between these two figures is modest, indicating that automated transcription approaches the level of agreement observed between independent human transcribers of the same audio. Errors were dominated by substitutions in most strata.

Transcription quality was consistent across the major structural dimensions of the corpus. WER differed only slightly by frequency band (5.37\% for AM versus 5.02\% for FM) and by station ownership cluster.
Variation was somewhat larger across Census regions, ranging from 4.17\% in the West to 6.91\% in the Northeast, though the
smaller regional subsamples yield correspondingly wider confidence intervals.
By programming format, WER was lowest for the more scripted or produced content types, such as religious service/sermon (3.87\%) and audio drama/narrative (3.54\%), and highest for spontaneous or acoustically difficult content, such as caller
interaction/audience participation (7.55\%) and transition/filler (8.08\%); the latter is estimated from only a small number of clips.
This pattern is consistent with the expectation that clear, planned speech is transcribed more accurately than overlapping, telephone-quality, or highly variable audio.

To characterize how transcription quality depends on recording conditions, we additionally labeled each validation clip with an automated audio-quality proxy computed directly from the waveform.
This proxy applies a set of simple heuristic checks to each clip: it flags recordings with a low proportion of detected speech, a low speech energy level, a low signal-to-noise proxy (the difference between speech and non-speech frame energy), a high fraction of clipped samples, or an excessive fraction of long silent regions.
A clip triggering one or more of these rules was labeled ASR-challenging; all others were labeled ASR-normal.
This labeling is derived entirely from the audio signal and does not use the reference transcript or the ASR output, so it provides an independent proxy for recording quality.
Clips labeled ASR-challenging (18.9\% of the validation sample) had a substantially higher WER (7.30\%) than ASR-normal clips (4.65\%), confirming that the pipeline's errors are concentrated in acoustically degraded recordings rather than distributed uniformly.
Because the same conditions that trigger these rules are observable throughout the full corpus, this stratification gives users a means of anticipating which recordings are likely to carry higher transcription error.

\begin{table}[H]
    \caption{Transcription quality metrics (WER)}
\centering
\label{tab:wer-by-stratum}
\begin{threeparttable}
    \small
\begin{tabular}{lrrrrrrr}
\toprule
\textbf{Stratum} & \textbf{Clips} & \textbf{Hrs} & \textbf{WER\%} & \textbf{95\% CI} & \textbf{Sub\%} & \textbf{Del\%} & \textbf{Ins\%} \\
\midrule
\multicolumn{7}{l}{\textit{Reference}} \\
\quad Human--Human (inter-annotator) & 100 & 1.52 & 2.57 & — & — & — & — \\
\quad Human--ASR (overall)           & 599 & 9.26 & 5.14 & — & — & — & — \\
\midrule
\multicolumn{7}{l}{\textit{Band}} \\
\quad AM & 190 & 2.95 & 5.37 & {[4.40, 6.55]} & 1.76 & 2.09 & 1.51 \\
\quad FM & 409 & 6.32 & 5.02 & {[4.35, 5.86]} & 1.73 & 1.56 & 1.74 \\
\midrule
\multicolumn{7}{l}{\textit{Census Region}} \\
\quad Northeast & 76  & 1.18 & 6.91 & {[4.44, 10.00]} & 1.68 & 2.77 & 2.46 \\
\quad South     & 252 & 3.90 & 5.43 & {[4.64, 6.28]}  & 1.95 & 1.74 & 1.74 \\
\quad Midwest   & 158 & 2.43 & 4.52 & {[3.55, 5.65]}  & 1.65 & 1.43 & 1.45 \\
\quad West      & 113 & 1.75 & 4.17 & {[3.16, 5.38]}  & 1.41 & 1.47 & 1.29 \\
\midrule
\multicolumn{7}{l}{\textit{Ownership Cluster (stations)}} \\
\quad Independent (1)       & 185 & 2.87 & 5.26 & {[4.17, 6.51]} & 1.80 & 1.50 & 1.95 \\
\quad Small (2--10)         & 246 & 3.79 & 4.88 & {[4.04, 5.93]} & 1.60 & 1.81 & 1.47 \\
\quad Medium (11--50)       &  97 & 1.50 & 5.99 & {[4.62, 7.53]} & 2.02 & 2.11 & 1.85 \\
\quad Large ($>$50)         &  71 & 1.10 & 4.58 & {[3.49, 5.85]} & 1.68 & 1.54 & 1.37 \\
\midrule
\multicolumn{7}{l}{\textit{Audio Quality}} \\
\quad ASR-challenging & 113 & 1.73 & 7.30 & {[5.41, 9.54]} & 2.15 & 2.72 & 2.42 \\
\quad ASR-normal      & 486 & 7.53 & 4.65 & {[4.10, 5.28]} & 1.65 & 1.51 & 1.50 \\
\midrule
\multicolumn{7}{l}{\textit{Programming Format}} \\
\quad Transition/filler                        &   3 & 0.05 &  8.08 & {[5.67, 10.48]} & 3.29 & 0.90 & 3.89 \\
\quad Caller/audience interaction              &  42 & 0.66 &  7.55 & {[5.30, 10.21]} & 2.43 & 2.13 & 2.98 \\
\quad Ad/promotion                             &  82 & 1.24 &  6.06 & {[4.28, 8.36]}  & 1.56 & 2.72 & 1.78 \\
\quad Discussion/monologue/commentary          & 189 & 2.93 &  5.73 & {[4.51, 7.23]}  & 1.58 & 2.11 & 2.04 \\
\quad Interview                                &  39 & 0.62 &  5.23 & {[3.77, 7.20]}  & 1.85 & 1.79 & 1.60 \\
\quad News/traffic/weather                     &  45 & 0.68 &  4.11 & {[3.27, 5.13]}  & 2.30 & 0.86 & 0.95 \\
\quad Religious service/sermon                 & 175 & 2.72 &  3.87 & {[3.17, 4.72]}  & 1.64 & 1.14 & 1.09 \\
\quad Audio drama/narrative                    &  24 & 0.38 &  3.54 & {[2.45, 4.93]}  & 1.70 & 0.52 & 1.32 \\
\bottomrule
\end{tabular}
\begin{tablenotes}
    \footnotesize
    \item Notes: Human--ASR WER by stratum for the transcribed radio broadcast corpus.
WER components are substitution (Sub), deletion (Del), and insertion (Ins) rates.
95\% confidence intervals are bootstrapped. Overall Human--ASR and Human--Human
(inter-annotator) WER are shown for reference.
\end{tablenotes}
\end{threeparttable}
\end{table}

\subsection*{LLM-based annotation quality}

The recording-level metadata includes two sets of LLM-generated labels: one that describes the programming format and one that provides topic labels per segment (see Methods).
To assess the reliability of these annotations, we validated each classification task against human-coded reference labels.

For both tasks, three researchers independently hand-coded a random sample of transcript segments using the same codebooks supplied to the model, and a single reference label (or set of labels, for the multi-label topic task) was derived for each segment by resolving disagreements among the three coders with a Dawid--Skene aggregation model~\cite{dawid_skene}, which estimates latent true labels while accounting for individual coder error rates.
Model predictions were then compared against these human-derived reference labels.
The two tasks are evaluated on different samples and with slightly different metrics, as described below.

We report macro- and micro-averaged $F_1$ together with overall percent agreement, and, for individual labels attested by at least 25 segments in the validation sample, label-level $F_1$ and percent agreement (Tables~\ref{tab:format_validation} and \ref{tab:topic_validation}).
Labels with fewer than 25 validation segments are too sparse to score reliably and are omitted; their absence reflects low prevalence in the validation sample rather than any assessment of model performance on those categories.

Format classification (Table~\ref{tab:format_validation}) was validated on a sample of 640 segments and reached a macro-$F_1$ of 0.77 and overall agreement of 74\%.
Performance was strongest for formats with distinctive lexical or structural cues, such as news read/traffic/weather ($F_1 = 0.84$) and ad/promotion ($F_1 = 0.83$), and weakest for categories that may blend into others, such as discussion/monologue/commentary ($F_1 = 0.63$) and transition/filler ($F_1 = 0.60$).

\begin{table}[H]
\centering
\caption{Format classification quality metrics}
\label{tab:format_validation}
\begin{threeparttable}
    \small
\begin{tabular}{lcc}
\toprule
 & F1 score & \% agreement \\
\midrule
\multicolumn{3}{l}{\emph{Model:} GPT-4.1 \quad \emph{Validation data:} 640 transcript segments} \\
\midrule
\multicolumn{3}{l}{\textbf{Aggregate}} \\
Macro F1 & 0.77 & --- \\
Micro F1 & 0.74 & --- \\
Overall \% agreement & --- & 74 \\
\midrule
\multicolumn{3}{l}{\textbf{Format (mutually exclusive)}} \\
Ad/promotion & 0.83 & 92 \\
Audio drama/narrative & * & * \\
Caller interaction/audience participation & 0.75 & 97 \\
Discussion/monologue/commentary & 0.63 & 84 \\
Interview & * & * \\
News read/traffic/weather & 0.84 & 98 \\
Religious service/sermon & 0.81 & 95 \\
Transition/filler & 0.60 & 86 \\
\bottomrule
\end{tabular}
\begin{tablenotes}
    \footnotesize
    \item Notes: Segment-level agreement between GPT-4.1 and human annotators
on the mutually exclusive format codebook. * Not enough cases in the validation
sample to calculate the metric for the individual label ($n < 25$).
\end{tablenotes}
\end{threeparttable}
\end{table}

Topic classification (Table~\ref{tab:topic_validation}) was validated on a sample of 352 segments and reached a macro-$F_1$ of 0.71 and a micro-$F_1$ of 0.79.
Because a segment may carry several topic labels, we additionally report the share of segments for which the model and the human reference agreed on at least one label (83\%).
Agreement was highest for the most concrete and prevalent topic, religion ($F_1 = 0.93$), and lowest for lifestyle/advice/personal development ($F_1 = 0.46$).
Overall, these results indicate that the format and topic labels are reliable enough to support aggregate, corpus-level description, while individual segment-level labels—especially in the lower-scoring categories—should be treated as approximate.

\begin{table}[H]
\centering
\caption{Topic classification quality metrics}
\label{tab:topic_validation}
\begin{threeparttable}
    \small
\begin{tabular}{lcc}
\toprule
 & F1 score & \% agreement \\
\midrule
\multicolumn{3}{l}{\emph{Model:} GPT-4.1 \quad \emph{Validation data:} 352 transcript segments} \\
\midrule
\multicolumn{3}{l}{\textbf{Aggregate}} \\
Macro F1 & 0.71 & --- \\
Micro F1 & 0.79 & --- \\
Overall \% agreement & --- & 91 \\
Agreement on $\geq$1 label & --- & 83 \\
\midrule
\multicolumn{3}{l}{\textbf{Topic (multi-label)}} \\
Business/economics/finance & * & * \\
Entertainment/pop culture/sports & * & * \\
Family/parenting/education & 0.73 & 94 \\
Health/wellness & 0.84 & 98 \\
Lifestyle/advice/personal development & 0.46 & 84 \\
Politics/current events/social commentary & 0.71 & 92 \\
Religion & 0.93 & 91 \\
Science/technology & * & * \\
\bottomrule
\end{tabular}
\begin{tablenotes}
    \footnotesize
    \item Notes: Segment-level agreement between GPT-4.1 and human annotators
on the multi-label topic codebook. ``Agreement on $\geq$1 label'' is the share of
cases in which the model and annotators agreed on at least one topic label. * Not
enough cases in the validation sample to calculate the metric for the individual
label ($n < 25$).
\end{tablenotes}
\end{threeparttable}
\end{table}


\section*{Usage Notes}

The dataset is distributed as Apache Parquet files and can be read with any Parquet-aware tool.
The Hugging Face release is intended as the primary access layer for programmatic use.
Because the \texttt{transcript\_lines} table is large, users may wish to stream it or load it in smaller segments rather than materializing the full corpus in memory.
The following examples use the \texttt{datasets} library.

\begin{lstlisting}[language=Python,basicstyle=\ttfamily\small,breaklines=true]
from datasets import load_dataset

# Station and recording metadata (each a single file)
stations = load_dataset("pew-data-labs/religious-radio-corpus",
                        data_files="stations/stations.parquet", 
                        split="train")
recordings = load_dataset("pew-data-labs/religious-radio-corpus",
                          data_files="recordings/recordings.parquet", 
                          split="train")

# Transcript lines for a single day (recommended)
day = load_dataset("pew-data-labs/religious-radio-corpus",
                   data_files="transcript_lines/2025-07-01/*.parquet",
                   split="train", streaming=True)

# Full transcript corpus via streaming
full = load_dataset("pew-data-labs/religious-radio-corpus",
                    data_files="transcript_lines/**/*.parquet",
                    split="train", streaming=True)
\end{lstlisting}

The three tables are linked by stable identifiers.
Transcript lines join to their recording on \texttt{recording\_id}, and recordings join to their stream entry on \texttt{station\_id}, allowing transcript text to be associated with recording-level format and topic labels and with stream-level metadata such as band and region.

\subsection*{Stream- versus station-level attribution}

Each row of the \texttt{stations} table, and each recording, corresponds to a monitored \emph{stream URL} rather than an individual licensed station.
Because stations under common ownership frequently rebroadcast a single shared feed, one stream may be carried by several stations.
Analyses reported at the station level must therefore expand each stream to its constituent stations using the
\texttt{related\_stations} field or the crosswalk table; doing so attributes the same recorded content to every station sharing the feed.
Users should confirm whether a given count refers to streams or to stations before comparing figures.

\subsection*{Speaker identifiers}

Speaker labels in \texttt{speaker\_id\_local} (\texttt{SPK\_00}, \texttt{SPK\_01}, \ldots) are assigned by the diarization pipeline independently within each recording.
They are not stable across recordings: the same label in two recordings does not denote the same person, and no cross-recording speaker linkage is provided.
Speaker-level analysis is therefore well defined only within a single recording.

\subsection*{Music segments}

Regions classified as music are not transcribed; they are represented by the placeholder token \texttt{<MUSIC>} in both the \texttt{text} and \texttt{speaker\_id\_local} fields and flagged by \texttt{is\_music}.
Users interested only in spoken content can filter these rows out using \texttt{is\_music}, while users studying the balance of music and speech can retain them.

\subsection*{Times and time zones}

Recording start times (\texttt{recording\_start\_utc}) are recorded in UTC.
Analyses that depend on local broadcast time---for example, examining programming by time of day or day of week---should convert to station-local time using the \texttt{utc\_adjust} offset or the IANA \texttt{timezone} field in the \texttt{stations} table.

\subsection*{Provenance of derived fields}

Transcripts are the output of an automated speech-recognition and diarization pipeline, and the format and topic labels in the \texttt{recordings} table are generated by a large language model applied at the segment level and then aggregated to the recording.
Both are therefore subject to error, and their accuracy varies by content type, audio quality, and other factors.
Users should consult the Technical Validation section for measured error rates before relying on these fields for a given application.

\section*{Data Availability}

The Religious Radio Corpus is available through two repositories.
The archival version of record is deposited in Harvard Dataverse~\cite{religiousradio-dataverse}, and an
access-optimized version in analysis-ready Parquet format is available through Hugging Face Datasets at \texttt{pew-data-labs/religious-radio-corpus}~\cite{religiousradio-huggingface}.
Both releases contain the \texttt{stations}, \texttt{recordings}, and \texttt{transcript\_lines} tables along with the stream-to-station crosswalk, as described in the Data Records section. 
The Dataverse deposit also includes schema and codebook files, as well as release documentation.
The released corpus contains only derived data products; the raw audio recordings are not included due to copyright and licensing restrictions.

\section*{Code Availability}

The code used to collect, transcribe, and annotate the corpus---including the stream-recording scheduler and capture containers, the speech/music classification and WhisperX transcription and diarization pipeline, and the LLM-based segmentation and labeling scripts---will be made publicly available at \url{https://github.com/pewresearch/religious-radio-corpus}.
The release will document the software versions and model identifiers used in each stage (these are also specified in the Methods section).

\newpage

\section*{Author Contributions}
S. Bestvater managed the project, including the data collection; contributed to the development of the collection pipeline code; and drafted this manuscript.
S. Seets, A. Chapekis, and A. Lieb contributed to the development of the collection pipeline code and to data collection and processing.
S. Shah provided technical guidance and contributed to code review.
A. Smith provided strategic guidance for the project.
All authors reviewed and approved the manuscript.

\section*{Competing Interests}
The authors declare no competing interests.

\section*{Acknowledgements}
We thank Luxuan Wang, Sawyer Reed, Sofia Conway, and Devin Teehan for their assistance with data annotation and validation for this project.

\section*{Funding}
This work was supported by the Pew-Knight Initiative, a research program jointly funded by The Pew Charitable Trusts and the John S. and James L. Knight Foundation.

\newpage
\section*{Appendix}
\appendix
\setcounter{table}{0}
\renewcommand{\thetable}{A\arabic{table}}

\subsection*{LLM prompt templates}

\begin{lstlisting}[language={},basicstyle=\ttfamily\small,breaklines=true]
You are an expert AI assistant designed to help process and extract 
relevant information from radio transcripts.
In a moment, you will be provided with a diarized transcript of about
15 minutes of talk radio programming.
The lines are numbered, and each line in the transcript represents a
single utterance, with a corresponding speaker label.
You will be asked to perform transcript segmentation.

Here are detailed instructions for the task:

Transcript Segmentation:

Individual utterances in radio programming can often be grouped
together into segments that cover a main general topic.
Please review the transcript and give your best suggestion for where
this transcript could be broken up into topical segments.
For each segment you suggest, please provide the following
information:
- starting_line_number: The starting line number of the segment
- ending_line_number: The ending line number of the segment
- segment_format: The format of the segment (music, religious service/
sermon, audio drama/narrative, ad/promotion, news read/traffic/
weather, caller interaction/audience participation, discussion/
monologue/commentary, interview, transition/filler, other)
- segment_topics: A list containing the primary topic (or topics) of
the segment (religion, politics/current events/social commentary,
health/wellness, family/parenting/education, science/technology,
entertainment/pop culture/sports, business/economics/finance,
lifestyle/advice/personal development, other)

{TRANSCRIPT TEXT}

\end{lstlisting}

\subsection*{Codebooks}

\begin{table}[H]
\centering
\caption{Format codebook.\newline Each segment is assigned exactly one format label (mutually exclusive).}
\label{tab:format-codebook}
\small
\begin{tabular}{p{2.6in} p{3.6in}}
\toprule
Format label & Definition \\
\midrule
Ad/promotion & An advertisement or promotion of a product, service, or event. \\
Audio drama/narrative & A dramatized production or audio play. \\
Caller interaction/audience participation & A host or DJ interacting with listeners, including taking calls from the audience and reading audience mail or comments. Does not include interaction with program guests. \\
Discussion/monologue/commentary & One or more speakers discussing a topic or providing commentary. Includes typical ``talk radio'' content, banter, religious opinions, and dialogue between hosts or DJs. \\
Interview & An interview with a program guest. Does not include dialogue between station hosts or DJs. \\
News read/traffic/weather & Straightforward reads of the news, or local traffic and weather conditions. Long-form commentary about news or current events is categorized as discussion/monologue/commentary. \\
Religious service/sermon & A single speaker preaching or delivering a religious message. Includes prayer, Mass, or liturgy. \\
Transition/filler & Administrative or logistical announcements, including show introductions, outros, segment transitions, or station identification. \\
\bottomrule
\end{tabular}
\end{table}

\begin{table}[H]
\centering
\caption{Topic codebook.\newline A segment may carry any number of topic labels (not mutually exclusive).}
\label{tab:topic-codebook}
\small
\begin{tabular}{p{2.6in} p{3.6in}}
\toprule
Topic label & Definition \\
\midrule
Business/economics/finance & Discussion of economic and financial topics, including tax or tariff policies, personal financial advice, stock markets or investments, and cryptocurrency. \\
Entertainment/pop culture/sports & Discussion of pop culture topics including film, television, music, sports, and celebrities. \\
Family/parenting/education & Discussion of family-related topics including marriage, having children, child care, child-rearing, schools, and education standards. \\
Health/wellness & Discussion of wellness-related topics including health care, abortion, vaccination, medication, mental health, fitness, and exercise. \\
Lifestyle/advice/personal development & Direct and prescriptive recommendations for lifestyle, actions, or behaviors. \\
Politics/current events/social commentary & Discussion of or commentary on news or current events, local updates, politics and policy, and social issues. Does not include community updates or weather/traffic segments that do not touch on broader political or news topics. \\
Religion & Discussion of religious topics, including religious beliefs, scripture, and religious figures. \\
Science/technology & Discussion of science-related topics, including climate change and evolution. \\
\bottomrule
\end{tabular}
\end{table}

\end{document}